\documentclass[conference]{IEEEtran_ISCAS22}
\IEEEoverridecommandlockouts
\usepackage{cite}
\usepackage{amsmath,amssymb,amsfonts}
\usepackage{algorithmic}
\usepackage[ruled,vlined]{algorithm2e}
\usepackage{graphicx}
\usepackage{subcaption}
\usepackage{textcomp}
\usepackage{acronym}
\usepackage[dvipsnames]{xcolor}
\usepackage{float}
\usepackage{siunitx}
\usepackage{units}
\usepackage{tikz}
\usepackage{pgf}
\usepackage{multirow}
\usepackage{pgfplots}
\usepackage{epstopdf} 
\usepackage{placeins}		
\graphicspath{{./img/}}

\def\BibTeX{{\rm B\kern-.05em{\sc i\kern-.025em b}\kern-.08em
    T\kern-.1667em\lower.7ex\hbox{E}\kern-.125emX}}
    
\DeclareGraphicsExtensions{.jpg,.pdf,.mps,.png}
\graphicspath{{img/} {./}} 

\begin{document}
\title{\vspace{-0.5cm}Hardware calibrated learning to compensate heterogeneity in analog RRAM-based Spiking Neural Networks  
\vspace{-0.3cm}
\thanks{}
}
\acrodef{ADC}[ADC]{Analog to Digital Converter}
\acrodef{ADEXP}[AdExp-I\&F]{Adaptive-Exponential Integrate and Fire}
\acrodef{AER}[AER]{Address-Event Representation}
\acrodef{AEX}[AEX]{AER EXtension board}
\acrodef{AE}[AE]{Address-Event}
\acrodef{AFM}[AFM]{Atomic Force Microscope}
\acrodef{AGC}[AGC]{Automatic Gain Control}
\acrodef{AI}[AI]{Artificial Intelligence}
\acrodef{ALD}[ALD]{Atomic Layer Deposition}
\acrodef{AMDA}[AMDA]{AER Motherboard with D/A converters}
\acrodef{ANN}[ANN]{Artificial Neural Network}
\acrodef{API}[API]{Application Programming Interface}
\acrodef{ARM}[ARM]{Advanced RISC Machine}
\acrodef{ASIC}[ASIC]{Application Specific Integrated Circuit}
\acrodef{AdExp}[AdExp-IF]{Adaptive Exponential Integrate-and-Fire}
\acrodef{BCI}[BCI]{Brain-Computer-Interface}
\acrodef{BCM}[BCM]{Bienenstock-Cooper-Munro}
\acrodef{BD}[BD]{Bundled Data}
\acrodef{BEOL}[BEOL]{Back-end of Line}
\acrodef{BG}[BG]{Bias Generator}
\acrodef{BMI}[BMI]{Brain-Machince Interface}
\acrodef{BTB}[BTB]{band-to-band tunnelling}
\acrodef{BP}[BP]{Back-propagation}
\acrodef{BPTT}[BPTT]{Back-propagation Through Time}
\acrodef{CAD}[CAD]{Computer Aided Design}
\acrodef{CAM}[CAM]{Content Addressable Memory}
\acrodef{CAVIAR}[CAVIAR]{Convolution AER Vision Architecture for Real-Time}
\acrodef{CA}[CA]{Cortical Automaton}
\acrodef{CCN}[CCN]{Cooperative and Competitive Network}
\acrodef{CDR}[CDR]{Clock-Data Recovery}
\acrodef{CFC}[CFC]{Current to Frequency Converter}
\acrodef{CHP}[CHP]{Communicating Hardware Processes}
\acrodef{CMIM}[CMIM]{Metal-insulator-metal Capacitor}
\acrodef{CML}[CML]{Current Mode Logic}
\acrodef{CMP}[CMP]{Chemical Mechanical Polishing}
\acrodef{CMOL}[CMOL]{Hybrid CMOS nanoelectronic circuits}
\acrodef{CMOS}[CMOS]{Complementary Metal-Oxide-Semiconductor}
\acrodef{CNN}[CCN]{Convolutional Neural Network}
\acrodef{COTS}[COTS]{Commercial Off-The-Shelf}
\acrodef{CPG}[CPG]{Central Pattern Generator}
\acrodef{CPLD}[CPLD]{Complex Programmable Logic Device}
\acrodef{CPU}[CPU]{Central Processing Unit}
\acrodef{CSM}[CSM]{Cortical State Machine}
\acrodef{CSP}[CSP]{Constraint Satisfaction Problem}
\acrodef{CV}[CV]{Coefficient of Variation}
\acrodef{DAC}[DAC]{Digital to Analog Converter}
\acrodef{DAS}[DAS]{Dynamic Auditory Sensor}
\acrodef{DAVIS}[DAVIS]{Dynamic and Active Pixel Vision Sensor}
\acrodef{DBN}[DBN]{Deep Belief Network}
\acrodef{DBS}[DBS]{Deep-Brain Stimulation}
\acrodef{DFA}[DFA]{Deterministic Finite Automaton}
\acrodef{DIBL}[DIBL]{drain-induced-barrier-lowering}
\acrodef{DI}[DI]{delay insensitive}
\acrodef{DMA}[DMA]{Direct Memory Access}
\acrodef{DNF}[DNF]{Dynamic Neural Field}
\acrodef{DNN}[DNN]{Deep Neural Network}
\acrodef{DoF}[DoF]{Degrees of Freedom}
\acrodef{DPE}[DPE]{Dynamic Parameter Estimation}
\acrodef{DPI}[DPI]{Differential Pair Integrator}
\acrodef{DRRZ}[DR-RZ]{Dual-Rail Return-to-Zero}
\acrodef{DRAM}[DRAM]{Dynamic Random Access Memory}
\acrodef{DR}[DR]{Dual Rail}
\acrodef{DSP}[DSP]{Digital Signal Processor}
\acrodef{DVS}[DVS]{Dynamic Vision Sensor}
\acrodef{DYNAP}[DYNAP]{Dynamic Neuromorphic Asynchronous Processor}
\acrodef{EBL}[EBL]{Electron Beam Lithography}
\acrodef{ECoG}[ECoG]{Electrocorticography}
\acrodef{EDVAC}[EDVAC]{Electronic Discrete Variable Automatic Computer}
\acrodef{EEG}[EEG]{electroencephalography}
\acrodef{EIN}[EIN]{Excitatory-Inhibitory Network}
\acrodef{EM}[EM]{Expectation Maximization}
\acrodef{EPSC}[EPSC]{Excitatory Post-Synaptic Current}
\acrodef{EPSP}[EPSP]{Excitatory Post-Synaptic Potential}
\acrodef{ET}[ET]{Eligibility Trace}
\acrodef{EZ}[EZ]{Epileptogenic Zone}
\acrodef{FDSOI}[FDSOI]{Fully-Depleted Silicon on Insulator}
\acrodef{FEOL}[FEOL]{Front-end of Line}
\acrodef{FET}[FET]{Field-Effect Transistor}
\acrodef{FFT}[FFT]{Fast Fourier Transform}
\acrodef{FI}[F-I]{Frequency-Current}
\acrodef{FPGA}[FPGA]{Field Programmable Gate Array}
\acrodef{FR}[FR]{Fast Ripple}
\acrodef{FSA}[FSA]{Finite State Automaton}
\acrodef{FSM}[FSM]{Finite State Machine}
\acrodef{GIDL}[GIDL]{gate-induced-drain-leakage}
\acrodef{GOPS}[GOPS]{Giga-Operations per Second}
\acrodef{GPU}[GPU]{Graphical Processing Unit}
\acrodef{GUI}[GUI]{Graphical User Interface}
\acrodef{HAL}[HAL]{Hardware Abstraction Layer}
\acrodef{HFO}[HFO]{High Frequency Oscillation}
\acrodef{HH}[H\&H]{Hodgkin \& Huxley}
\acrodef{HMM}[HMM]{Hidden Markov Model}
\acrodef{HRS}[HRS]{High-Resistive State}
\acrodef{HR}[HR]{Human Readable}
\acrodef{HSE}[HSE]{Handshaking Expansion}
\acrodef{HW}[HW]{Hardware}
\acrodef{IBCI}[IBCI]{Implantable BCI}
\acrodef{ICT}[ICT]{Information and Communication Technology}
\acrodef{IC}[IC]{Integrated Circuit}
\acrodef{ICL}[ICL]{Implantable Closed Loop}
\acrodef{IEEG}[iEEG]{intracranial electroencephalography}
\acrodef{IF2DWTA}[IF2DWTA]{Integrate \& Fire 2--Dimensional WTA}
\acrodef{IFSLWTA}[IFSLWTA]{Integrate \& Fire Stop Learning WTA}
\acrodef{IF}[I\&F]{Integrate-and-Fire}
\acrodef{IMU}[IMU]{Inertial Measurement Unit}
\acrodef{INCF}[INCF]{International Neuroinformatics Coordinating Facility}
\acrodef{INI}[INI]{Institute of Neuroinformatics}
\acrodef{INRC}[Intel NRC]{Intel Neuromorphic Research Community}
\acrodef{IO}[I/O]{Input/Output}
\acrodef{IoT}[IoT]{Internet of Things}
\acrodef{IPSC}[IPSC]{Inhibitory Post-Synaptic Current}
\acrodef{IPSP}[IPSP]{Inhibitory Post-Synaptic Potential}
\acrodef{IP}[IP]{Intellectual Property}
\acrodef{ISI}[ISI]{Inter-Spike Interval}
\acrodef{IoT}[IoT]{Internet of Things}
\acrodef{JFLAP}[JFLAP]{Java - Formal Languages and Automata Package}
\acrodef{LEDR}[LEDR]{Level-Encoded Dual-Rail}
\acrodef{LFP}[LFP]{Local Field Potential}
\acrodef{LLC}[LLC]{Low Leakage Cell}
\acrodef{LNA}[LNA]{Low-Noise Amplifier}
\acrodef{LPF}[LPF]{Low Pass Filter}
\acrodef{LRS}[LRS]{Low-Resistive State}
\acrodef{LSM}[LSM]{Liquid State Machine}
\acrodef{LTD}[LTD]{Long Term Depression}
\acrodef{LTI}[LTI]{Linear Time-Invariant}
\acrodef{LTP}[LTP]{Long Term Potentiation}
\acrodef{LTU}[LTU]{Linear Threshold Unit}
\acrodef{LUT}[LUT]{Look-Up Table}
\acrodef{LVDS}[LVDS]{Low Voltage Differential Signaling}
\acrodef{MD}[MD]{Medical Device}
\acrodef{MCMC}[MCMC]{Markov-Chain Monte Carlo}
\acrodef{MEMS}[MEMS]{Micro Electro Mechanical System}
\acrodef{MFR}[MFR]{Mean Firing Rate}
\acrodef{MIM}[MIM]{Metal Insulator Metal}
\acrodef{ML}[ML]{Machine Leanring}
\acrodef{MLP}[MLP]{Multilayer Perceptron}
\acrodef{MOSCAP}[MOSCAP]{Metal Oxide Semiconductor Capacitor}
\acrodef{MOSFET}[MOSFET]{Metal Oxide Semiconductor Field-Effect Transistor}
\acrodef{MOS}[MOS]{Metal Oxide Semiconductor}
\acrodef{MRI}[MRI]{Magnetic Resonance Imaging}
\acrodef{NDFSM}[NDFSM]{Non-deterministic Finite State Machine} 
\acrodef{ND}[ND]{Noise-Driven}
\acrodef{NEF}[NEF]{Neural Engineering Framework}
\acrodef{NHML}[NHML]{Neuromorphic Hardware Mark-up Language}
\acrodef{NIL}[NIL]{Nano-Imprint Lithography}
\acrodef{NLP}[NLP]{Natural Language Processing}
\acrodef{NMDA}[NMDA]{N-Methyl-D-Aspartate}
\acrodef{NME}[NE]{Neuromorphic Engineering}
\acrodef{NN}[NN]{Neural Network}
\acrodef{NRZ}[NRZ]{Non-Return-to-Zero}
\acrodef{NSM}[NSM]{Neural State Machine}
\acrodef{OR}[OR]{Operating Room}
\acrodef{OTA}[OTA]{Operational Transconductance Amplifier}
\acrodef{PCB}[PCB]{Printed Circuit Board}
\acrodef{PCHB}[PCHB]{Pre-Charge Half-Buffer}
\acrodef{PCM}[PCM]{Phase Change Memory}
\acrodef{PD}[PD]{Parkinson Disease}
\acrodef{PE}[PE]{Phase Encoding}
\acrodef{PFA}[PFA]{Probabilistic Finite Automaton}
\acrodef{PFC}[PFC]{prefrontal cortex}
\acrodef{PFM}[PFM]{Pulse Frequency Modulation}
\acrodef{PM}[PM]{Personalized Medicine}
\acrodef{PR}[PR]{Production Rule}
\acrodef{PSC}[PSC]{Post-Synaptic Current}
\acrodef{PSP}[PSP]{Post-Synaptic Potential}
\acrodef{PSTH}[PSTH]{Peri-Stimulus Time Histogram}
\acrodef{PVD}[PVD]{Physical Vapor Deposition }
\acrodef{QDI}[QDI]{Quasi Delay Insensitive}
\acrodef{RAM}[RAM]{Random Access Memory}
\acrodef{RDF}[RDF]{random dopant fluctuation}
\acrodef{RELU}[ReLu]{Rectified Linear Unit}
\acrodef{RLS}[RLS]{Recursive Least-Squares}
\acrodef{RMSE}[RMSE]{Root Mean Squared-Error}
\acrodef{RMS}[RMS]{Root Mean Squared}
\acrodef{RNN}[RNN]{Recurrent Neural Network}
\acrodef{ROLLS}[ROLLS]{Reconfigurable On-Line Learning Spiking}
\acrodef{RRAM}[R-RAM]{Resistive Random Access Memory}
\acrodef{R}[R]{Ripples}
\acrodef{SAC}[SAC]{Selective Attention Chip}
\acrodef{SAT}[SAT]{Boolean Satisfiability Problem}
\acrodef{SCX}[SCX]{Silicon CorteX}
\acrodef{SD}[SD]{Signal-Driven}
\acrodef{SDSP}[SDSP]{Spike Driven Synaptic Plasticity}
\acrodef{SEM}[SEM]{Spike-based Expectation Maximization}
\acrodef{SLAM}[SLAM]{Simultaneous Localization and Mapping}
\acrodef{SNN}[SNN]{Spiking Neural Network}
\acrodef{SNR}[SNR]{Signal to Noise Ratio}
\acrodef{SOC}[SOC]{System-On-Chip}
\acrodef{SOI}[SOI]{Silicon on Insulator}
\acrodef{SoA}[SoA]{state-of-the-art}
\acrodef{SP}[SP]{Separation Property}
\acrodef{SRAM}[SRAM]{Static Random Access Memory}
\acrodef{STDP}[STDP]{Spike-Timing Dependent Plasticity}
\acrodef{STD}[STD]{Short-Term Depression}
\acrodef{STEM}[STEM]{Science, Technology, Engineering and Mathematics}
\acrodef{STP}[STP]{Short-Term Plasticity}
\acrodef{STT-MRAM}[STT-MRAM]{Spin-Transfer Torque Magnetic Random Access Memory}
\acrodef{STT}[STT]{Spin-Transfer Torque}
\acrodef{SW}[SW]{Software}
\acrodef{TCAM}[TCAM]{Ternary Content-Addressable Memory}
\acrodef{TFT}[TFT]{Thin Film Transistor}
\acrodef{TPU}[TPU]{Tensor Processing Unit}
\acrodef{TRL}[TRL]{Technology Readiness Level}
\acrodef{USB}[USB]{Universal Serial Bus}
\acrodef{VHDL}[VHDL]{VHSIC Hardware Description Language}
\acrodef{VLSI}[VLSI]{Very Large Scale Integration}
\acrodef{VOR}[VOR]{Vestibulo-Ocular Reflex}
\acrodef{WCST}[WCST]{Wisconsin Card Sorting Test}
\acrodef{WTA}[WTA]{Winner-Take-All}
\acrodef{XML}[XML]{eXtensible Mark-up Language}
\acrodef{CTXCTL}[CTXCTL]{CortexControl}
\acrodef{divmod3}[DIVMOD3]{divisibility of a number by three}
\acrodef{hWTA}[hWTA]{hard Winner-Take-All}
\acrodef{sWTA}[sWTA]{soft Winner-Take-All}
\acrodef{APMOM}[APMOM]{Alternate Polarity Metal On Metal}
 \author{\IEEEauthorblockN{
 Filippo Moro\textsuperscript{1}, 
 E. Esmanhotto\textsuperscript{1},
 T. Hirtzlin\textsuperscript{1},
 N. Castellani\textsuperscript{1},
 A. Trabelsi\textsuperscript{1},
 T. Dalgaty\textsuperscript{1},
 G. Molas\textsuperscript{1},\\
 F. Andrieu\textsuperscript{1},
 S. Brivio\textsuperscript{2},
 S. Spiga\textsuperscript{2},
 G. Indiveri\textsuperscript{3},
 M. Payvand\textsuperscript{3}, and
 E. Vianello\textsuperscript{1}}
 \IEEEauthorblockA{
 \textsuperscript{1}\textit{\ CEA-Leti, Grenoble, France},
 \textsuperscript{2}\textit{\ CNR-IMM, Agrate Brianza, Italy} \\
 \textsuperscript{3}\textit{\ Institute of Neuroinformatics, University of Zurich and ETH Zurich, Switzerland}\\
 \vspace{-1.3cm}}

 }
\maketitle
\begin{abstract}
  Spiking Neural Networks (SNNs) can unleash the full power of analog Resistive Random Access Memories (RRAMs) based circuits for low power signal processing.
  Their inherent computational sparsity naturally results in energy efficiency benefits.
  The main challenge implementing robust SNNs is the intrinsic variability (heterogeneity) of both analog CMOS circuits and RRAM technology.
  In this work, we assessed the performance and variability of RRAM-based neuromorphic circuits that were designed and fabricated using a 130\,nm technology node.
  Based on these results, we propose a Neuromorphic Hardware Calibrated (NHC) SNN, where the learning circuits are calibrated on the measured data.
  We show that by taking into account the measured heterogeneity characteristics in the off-chip learning phase, the NHC SNN self-corrects its hardware non-idealities and learns to solve benchmark tasks with high accuracy.
  This work demonstrates how to cope with the heterogeneity of neurons and synapses for increasing classification accuracy in temporal tasks.
\end{abstract}

\vspace*{-1mm}
\section{Introduction}
\begin{figure}[t]
    \vspace*{-0.4cm}
	\centering
    \includegraphics[width=0.38\textwidth]{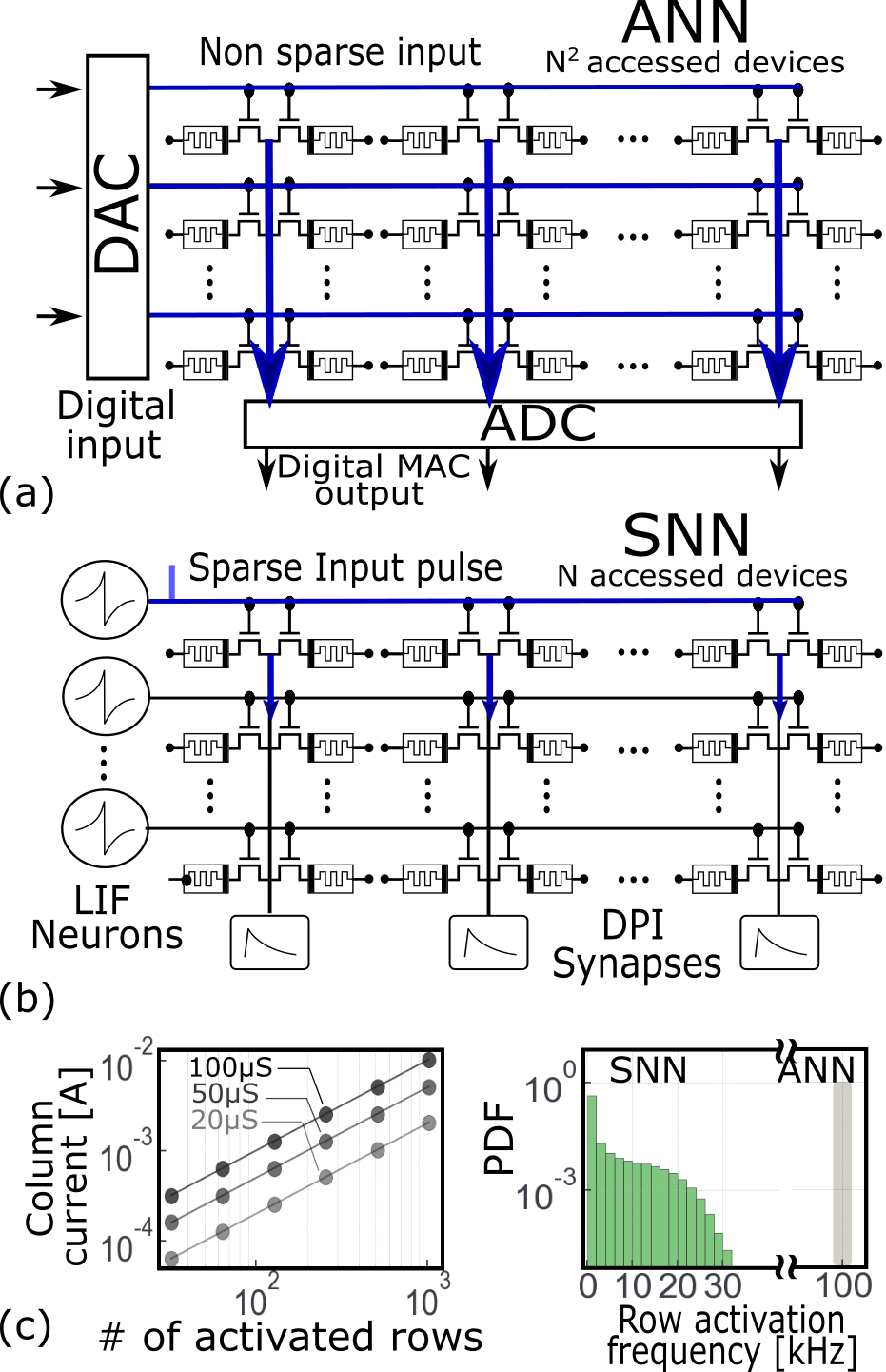}
    \caption{RRAM crossbar arrays in ANN (a) and SNN (b). (c) Quantification of current magnitude per column for different average RRAM conductance and row activation frequency distribution for ANN and SNN.}
    \label{fig:ANNvsSNN}
    \FloatBarrier
\end{figure}

Resistive Random Access Memories (RRAMs) have been shown to have a large potential for locally storing the synaptic weights and enabling ``in memory computing'' in Artificial Neural Networks (ANNs)~\cite{Sebastian_2020}.
The assembly of resistive memories organized as a crossbar naturally implements the Multiply And Accumulate (MAC) operation in ANNs (Fig.~\ref{fig:ANNvsSNN}).
However, one of the major problems of this ANN approach is network scalability.
In this approach, the output current at each column, and the overall power budget increase linearly with the number of devices being read (i.e. number of activated rows), thus strongly limiting the array size (Fig.~\ref{fig:ANNvsSNN}).
Another limitation is the overhead required by the Digital-to-Analog (DAC) and Analog-to-Digital (ADC) circuits needed for the conversion.
To overcome these issues we focus on the hardware implementation of analog Spiking Neural Networks (SNNs).
SNNs have typically very sparse activations, so the number of activated rows at any instance of time is very small, significantly reducing the current and power consumption at each column (Fig.~\ref{fig:ANNvsSNN}). Moreover, analog neurons and synapses in SNNs do not require DACs and ADCs, resulting in a further reduction of energy consumption and area~\cite{Valentinian_2019}.

The basic building block of analog SNNs is composed of Leaky Integrate and Fire (LIF) neurons. In SNNs LIF neurons transmit voltage pulses (spikes) to multiple columns of one resistor-one-transistor (1T1R) devices, which encode the network synaptic weights in their conductance. The resulting current is the weighted sum of all the synaptic outputs. In the architecture we propose these currents are then integrated temporally by a shared Differential Pair Integrator (DPI) circuit (Fig.~\ref{fig:ANNvsSNN}), which is a subthreshold log-domain low-pass filter~\cite{DPI_neuron_Livi_2009, Bartolozzi_2007}. To realize a multi-layer neural network, such basic blocks can be chained together in a modular way.

However, both analog circuits and RRAMs exhibit device variability. In this work, we compared the performance of SNNs trained to carry out three different tasks, with different degrees of hardware heterogeneity. The CMOS variability affects the neuron's and synapse's time constants. The RRAM variability affects the synaptic weights. We propose a Neuromorphic Hardware Calibrated (NHC) SNN, where off-chip training is calibrated on experimentally measured data and hardware non-idealities. This approach allows achieving classification accuracy on three different tasks, comparable to equivalent full-precision (32-bit floating point) software-based simulations. Moreover, we demonstrated that heterogeneity in neuron and synapse time constants originates a richer system temporal dynamics, thus improving the accuracy for tasks with temporal structure. Experiments have been conducted on custom analog LIF neuron and DPI synapse circuits as well as on a 4~kb HfO2 crossbar 1T1R memory array fabricated in a commercial 130\,nm technology node.

\vspace{-0.2cm}
\section{Heterogeneity in neurons and synapses}
\begin{figure}[t]
	\centering
    \includegraphics[width=0.48\textwidth]{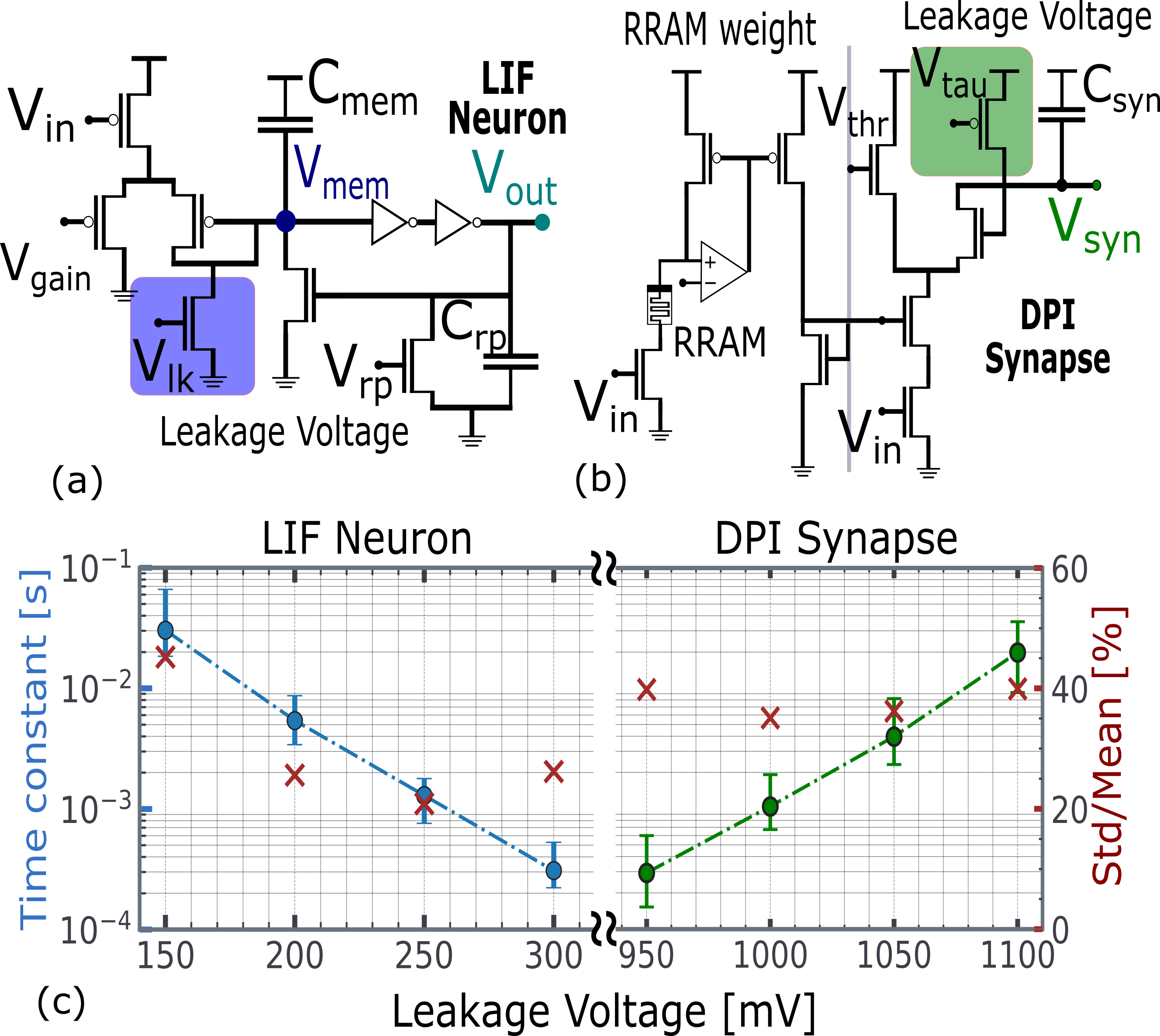}
    \caption{LIF neuron (a) and DPI-RRAM synapse (b) circuits. (c) Time constants in the neuron and synapse circuits as a function of the biase leak voltage.}
    \label{fig:time_constant_variability}
\end{figure}

We designed, fabricated, and tested analog CMOS-based LIF neuron and synapse circuits (Fig.~\ref{fig:time_constant_variability}a,b). The design is based on the DPI circuit \cite{DPI_neuron_Livi_2009, Bartolozzi_2007}, which implements a low pass filter with time constant controlled by a tunable bias voltage. In arrays of such circuits the same bias voltage produces heterogeneous leak current.
To derive the DPI circuit time constant we applied an input voltage pulse at the input (Vin) and measured the voltage at the capacitor. The resulting trace was fitted with an exponential function. By modulating the Vlk bias of the neuron (or Vtau biase of the synapse), we modified the current leak rate, resulting in different time constants (Fig.~\ref{fig:time_constant_variability}c). The measurements have been repeated over 100 samples and the time constant extrapolated from the response of Vmem/Vsyn. Variability in the neuron and synapse time constants is quantified at about 30\% in standard deviation over the mean.

\section{Variability in the RRAM and synaptic weights}
\begin{figure}
	\centering
    \includegraphics[width=0.49\textwidth]{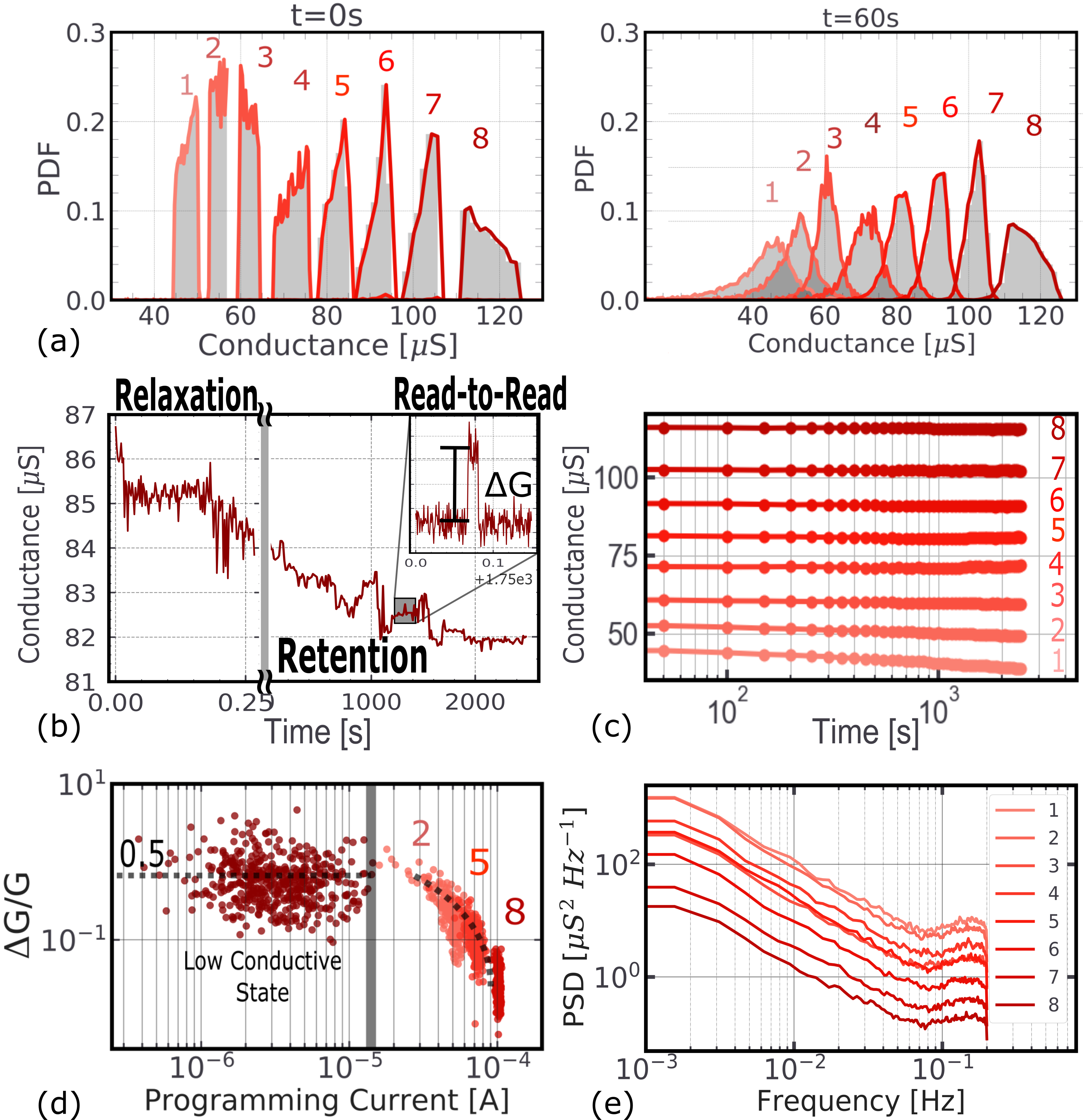}
    \caption{(a) Multilevel programming of 8 conductance levels at t=0\,s and t=60\,s. (b) Temporal variability effects after programming. (c) Level distribution mean, measured over time. (d) Measured $\Delta G/ G$ as a function of the programming current ($\Delta G$ is due to RTN and is defined in (b)). (e) Power Spectral density of the noise in the 8 conductance levels. 
    }
    \label{fig:rrams}
    \vspace{-0.2cm}
\end{figure}

To obtain 8 conductance levels per RRAM device in a 4~kb 1T1R array, we used the multilevel smart programming procedure described in~\cite{Esmanhotto_2020}. We then measured and characterized the distribution of the conductances in time (see Fig.~\ref{fig:rrams}a). As shown, the smart programming procedure yields tightly distributed conductance levels, which broaden with time due to temporal variability of the devices. RRAMs show 3 degrees of temporal variability that take place at different time scales (Fig.~\ref{fig:rrams}b). Relaxation takes place just after programming (milliseconds) and broadens all the conductance levels distributions. Data retention causes long term (hours) variation of the conductance, particularly affecting the lower conductance levels, whose mean of the distribution decreases with time (Fig.~\ref{fig:rrams}c). Read-to-Read (R2R) noise does not affect the shape of the conductance distribution, although when looking at individual devices there are fast temporal fluctuations of conductance due to reading disturbances and Random Telegraph Noise (RTN). We evaluate the RTN component in R2R via the $\Delta G/G$ figure of merit (Fig.~\ref{fig:rrams}d), measuring the conductance jumps $\Delta G$ due to RTN. The result is in line with the literature \cite{Puglisi_2016}. Finally, the Power Spectral Density of the 8 conductance levels shows that the amount of noise is inversely proportional to the conductance and is general of the 1/f type (Fig.~\ref{fig:rrams}e), as also observed in \cite{Ambrogio_2015}.

\section{Hardware-calibrated off-chip learning}
We trained the SNN off-chip with the Surrogate Gradient algorithm~\cite{Neftci_2019}, using 32-bits floating point weights: this technique allows to take into account the non-idealities of the hardware substrate in the learning phase. Heterogeneity is introduced by assigning each neuron and synapse a different time constant value sampled from the experimental distributions of Fig.~\ref{fig:time_constant_variability}c. The procedure is completed by transferring the learned weights to the RRAM array, by   discretizing them to 3-bit values and converting them to the corresponding conductance levels. As the training accounts for the variability of both analog circuits and RRAM devices, we defined it as Neuromorphic Hardware Calibrated (NHC) procedure. This procedure is applied to three different benchmark tasks with different degrees of temporal structure: MNIST (static visual image of handwritten digits), ECG \cite{ECG_Moody_2001} (heart arrhythmia classification), and SHD \cite{SHD_Cramer_2020} (spoken digits). In all the cases the architecture of the network features 128 neurons in the hidden layer, with recurrent connections enabled for the ECG and SHD tasks. Input and Output layer dimensions depend on the task. For the ECG case, the 5 most frequent heart diseases in the dataset are selected for classification.

\begin{figure}
	\centering
    \includegraphics[width=0.42\textwidth]{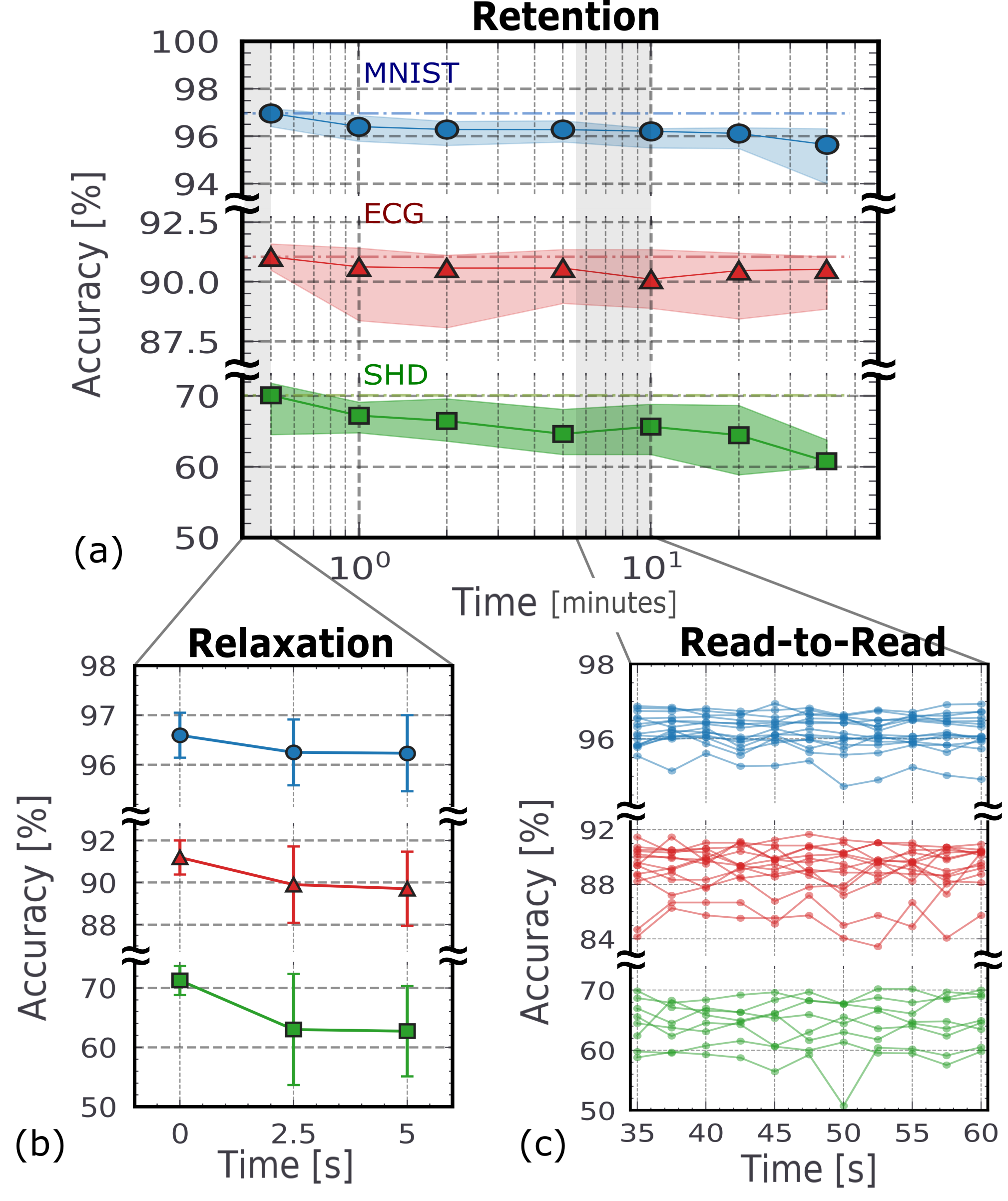}
    \caption{Accuracy for the three benchmark tasks, tested with the RRAM array measured across time. (a) Data Retention acts over the course of hours, reducing accuracy. (b) Relaxation induces an accuracy drop after programming. (c) R2R causes small variations of conductance each time RRAM are read, slightly perturbing performance. }
    \label{fig:results}
\end{figure}

\section{Impact of heterogeneity on performance}
In Table \ref{tab:results} we list the effect of the measured analog circuits heterogeneity on the performance of the network, compared to the case of ideal SNNs (Homogeneous SNN) and software-based ANNs. Ignoring hardware heterogeneity in the training phase (Non-Calibrated SNN) and then performing inference on a heterogeneous hardware network causes the accuracy of the SNN to drop by more than 10\%. The proposed NHC training approach recovers this loss in performances. Moreover, heterogeneity in time constants surprisingly improves the accuracy on datasets with rich temporal structure (ECG and SHD). This result is in agreement with the theoretical study performed in \cite{PerezNieves_2020} and could be explained by the richer temporal dynamics of the heterogeneous substrate.

\begin{table}[h]
\resizebox{0.48\textwidth}{!}{%
\begin{tabular}{ccccc}
\hline
                                 &Weights                &N-MNIST      &ECG              &SHD              \\ \hline
ANN                              &float32        &97.5\%       &95.5\%  &89.0\% \\ \hline
NC SNN                    &float32        &90.2\%       &63.7\%           &58.4\%           \\ \hline
\multirow{2}{*}{Hom. SNN} &float32        &97.4\%       &94.5\%           &72.5\%           \\
                                 &4bits      &96.7\%       &91.4\%           &71.6\%           \\ \hline
\multirow{5}{*}{NHC SNN}         &float32    &\textbf{97.5\%}       &\textbf{94.9\%}           &\textbf{74.9\%}           \\
                                 &4bits      &96.9\%       &91.4\%           &73.2\%           \\
                                 &RRAM t=0s  &96.8\%       &91.2\%           &71.2\%           \\
                                 &RRAM t=5s  &96.2\%       &90.2\%           &67.5\%           \\
                                 &RRAM t=1h  &95.3\%       &89.9\%           &60.4\%           \\ \hline
\end{tabular}%
}
\caption{NHC SNN results and comparison with ANN, Non-Calibrated SNN (NC SNN) and Homogenous SNN (Hom. SNN).}
\label{tab:results}
\end{table}

\section{Impact of RRAM non-idealities on performance}
\begin{figure}[t]
	\centering
    \includegraphics[width=0.36\textwidth]{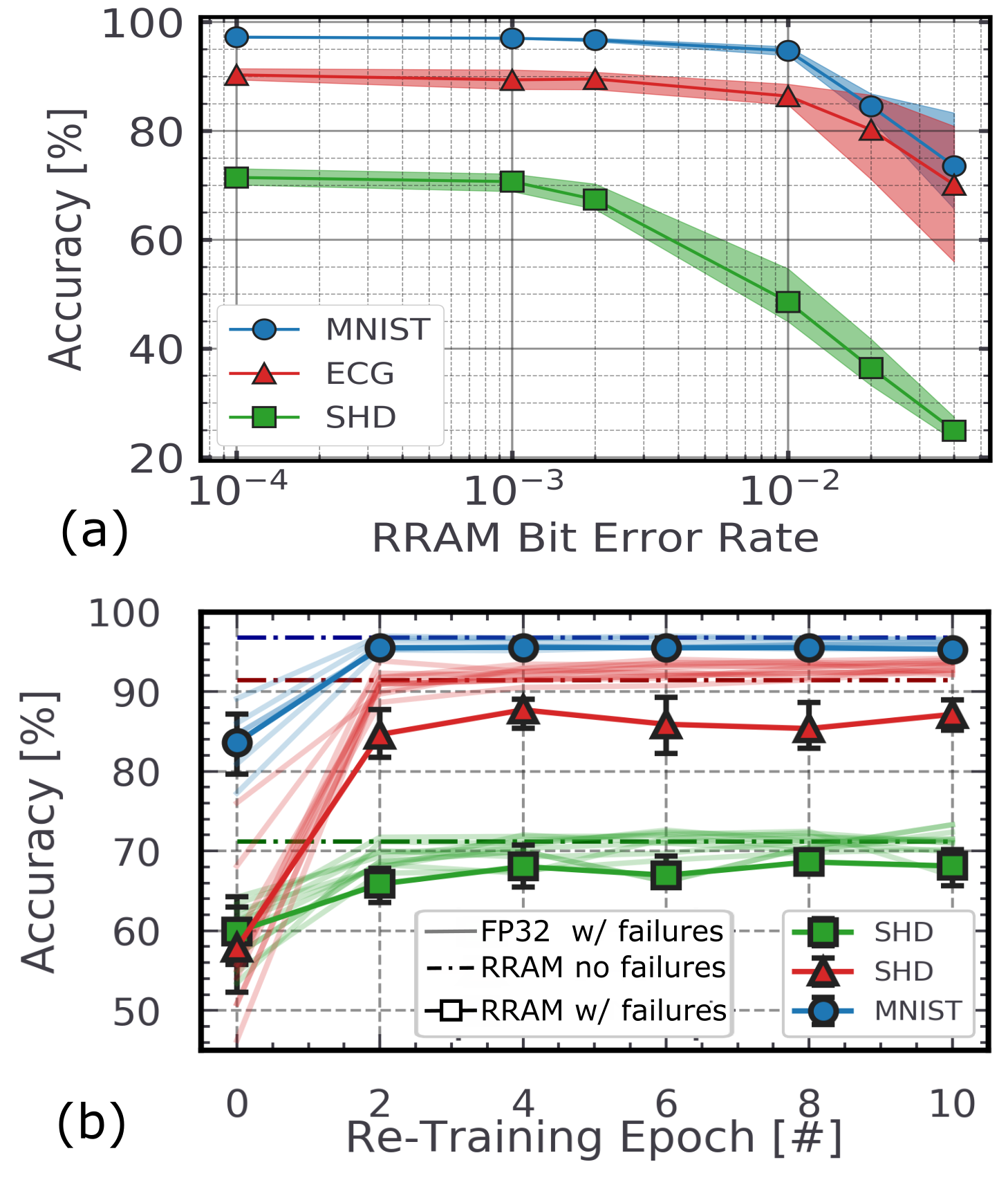}
    \caption{Analysis of performance with RRAM failures and re-training taking the failures into account. (a) Accuracy as a function of the Bit-Error-Rate (BER) of RRAM weights. (b) Networks with high degree of RRAM failures (BER or $10^{-2}$) re-trained considering the weight defects.}
    \label{fig:retraining}
\end{figure}

The RRAMs support up to 8 distinct conductance levels, enough to saturate performance for simple datasets, as demonstrated in \cite{Esmanhotto_2020}. The impact of the RRAM temporal variability is shown in Fig~\ref{fig:results}. Relaxation causes an immediate decrease in performance (Fig~\ref{fig:results}b). The decrease of performance over time due to poor data retention (Fig~\ref{fig:results}a) is minimal for simpler tasks like MNIST (blue) and ECG (red), while it is more pronounced for SHD (green). R2R noise slightly varies the conductance values at each inference operation (Fig~\ref{fig:results}c), causing accuracy to fluctuate. Furthermore, the impact of failures in the RRAM-based neuromorphic chip is evaluated. A failure is represented by a device stuck at either low ($1\mu S \pm 0.5\mu S$) or high ($200\mu S \pm 25\mu S$) conductance. The accuracy as a function of the RRAM’s Bit Error Rate (BER) is shown in Fig~\ref{fig:retraining}a: SNN models are resilient up to BER of $10^{-3}$. In order to mitigate faults, we can retrain the SNN with broken RRAMs (Fig~\ref{fig:retraining}b), to recover performance. MNIST is re-learned with just one learning epoch, while ECG and SHD require a few more epochs to recover. Overall, the performance is almost fully restored in all cases.

\section{Energy assessment}
\begin{figure}[t]
	\centering
    \includegraphics[width=0.38\textwidth]{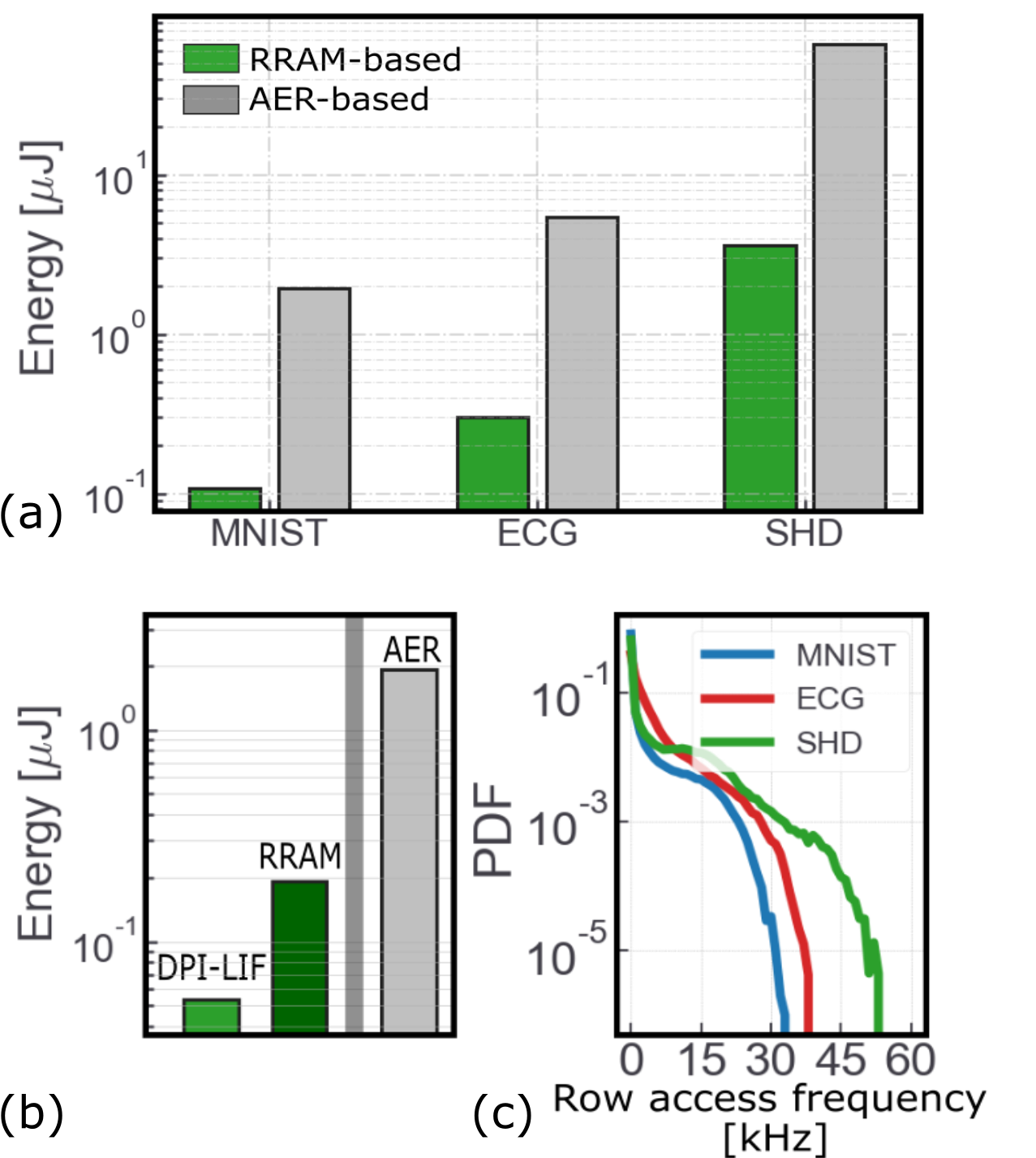}
    \caption{(a) Energy per inference step of RRAM-based SNN, compared to DYNAP\cite{DYNAPS_Moradi_2018}. (b) Energy contributions of the RRAMs and analog circuit (DPI-LIF), for the MNIST benchmark, compared to the routing in DYNAP. (c) Row access statistics show the sparse computation features of the SNN.}
    \label{fig:energy}
\end{figure}

To assess the efficiency of an RRAM based neuromorphic processor we compare their energy per inference sample with a mixed-signal neuromorphic processor, DYNAP \cite{DYNAPS_Moradi_2018}. DYNAP uses similar LIF neuron and DPI synapse circuits, but employs an asynchronous digital communication protocol to implement network connectivity. The energy consumption for the RRAM-based system is estimated by means of SPICE simulations and is more than 1 order of magnitude lower than that of DYNAP (Fig.~\ref{fig:energy}a). Energy is dominated by the RRAMs (that store the synaptic weights and define the network topology) in the reading operation. However, the RRAM associated energy is about 1 order of magnitude less than that of the communication protocol used in DYNAP (Fig.~\ref{fig:energy}b). Furthermore, SNN computation is very sparse, reducing the number of simultaneously activated rows of the RRAM array, yielding small currents on the column lines (Fig.~\ref{fig:energy}c).

\section{Conclusion}
We proposed a new approach for training RRAM-based analog SNN that takes into account the hardware details. The results show, that SNNs trained with our approach reach competitive classification accuracy levels, and that the heterogeneity of neurons and synapses improves network performance for temporal tasks. Although the use of RRAMs could result in slightly reduced performance over time, they can reduce the energy cost per inference by one order of magnitude with respect to conventional Mixed-Signal processors.

\section*{Acknowledgment}
This work is supported by the H2020 MeM-Scales project (871371) and the ANR Carnot funding.

\bibliographystyle{IEEEtran}
\bibliography{refs.bib}
\end{document}